%% file: root.tex
\documentclass[letterpaper, 10 pt, conference]{ieeeconf}  

\IEEEoverridecommandlockouts                              

\usepackage{graphicx}
\usepackage[ruled,vlined]{algorithm2e}
\usepackage{mathtools}

\usepackage{float} 
\usepackage{amsmath} 

\DeclareMathOperator*{\argmin}{arg\,min}

\usepackage{balance}
\usepackage{bbold}
\usepackage{bbm}
\usepackage{hyperref}
\usepackage{textcomp}
\usepackage{dblfloatfix}
\usepackage{booktabs}
\let\labelindent\relax
\usepackage{enumitem}

\title{\LARGE \bf
A Method For Automated Drone Viewpoints to Support\\ Remote Robot Manipulation
}

\author{Emmanuel Senft$^{1,\dagger}$, Michael Hagenow$^{2,\dagger}$,  Pragathi Praveena$^{1}$\\ Robert Radwin$^{3}$, Michael Zinn$^{2}$, Michael Gleicher$^{1}$, and Bilge Mutlu$^{1}$ 
\thanks{This work was supported in part by a NASA University Leadership Initiative (ULI) grant awarded to the UW-Madison and The Boeing Company (Cooperative Agreement \# 80NSSC19M0124).}
\thanks{$^{1}$Emmanuel Senft, Pragathi Praveena, Michael Gleicher, and Bilge Mutlu are with the Department of Computer Sciences, University of Wisconsin--Madison, Madison 53706, USA.
        {\tt\small [esenft|pragathi|gleicher|bilge]@cs.wisc.edu}}
\thanks{$^{2}$Michael Hagenow and Michael Zinn are with the Department of Mechanical Engineering, University of Wisconsin--Madison, Madison 53706, USA.
        {\tt\small [mhagenow|mzinn]@wisc.edu}}%
\thanks{$^{3}$Robert Radwin is with the Department of Industrial and Systems Engineering, University of Wisconsin--Madison, Madison 53706, USA.
        {\tt\small rradwin@wisc.edu}}%
\thanks{$^{\dagger}$These two authors contributed equally to the work, and consequently should share first authorship.}%
}

\begin{document}

\maketitle
\thispagestyle{empty}
\pagestyle{empty}

\begin{abstract}
Drones can provide a minimally-constrained adapting camera view to support robot telemanipulation. Furthermore, the drone view can be automated to reduce the burden on the operator during teleoperation. However, existing approaches do not focus on two important aspects of using a drone as an automated view provider. The first is how the drone should select from a range of quality viewpoints within the workspace (e.g., opposite sides of an object). The second is how to compensate for unavoidable drone pose uncertainty in determining the viewpoint. In this paper, we provide a nonlinear optimization method that yields effective and adaptive drone viewpoints for telemanipulation with an articulated manipulator. Our first key idea is to use sparse human-in-the-loop input to toggle between multiple automatically-generated drone viewpoints. Our second key idea is to introduce optimization objectives that maintain a view of the manipulator while considering drone uncertainty and the impact on viewpoint occlusion and environment collisions. We provide an instantiation of our drone viewpoint method within a drone-manipulator remote teleoperation system. Finally, we provide an initial validation of our method in tasks where we complete common household and industrial manipulations.

\end{abstract}

\input{1-introduction}

\input{2-related}
\input{3-method}

\input{4-implementation}

\input{5-case_study}

\input{6-discussion}




\bibliographystyle{IEEEtran}
\bibliography{root}
\balance

\end{document}

%% file: 1-introduction.tex
\section{INTRODUCTION} \label{sec:introduction}

Remotely teleoperated robots can assist in many unstructured and inaccessible settings. Effective telemanipulation in these environments requires the remote operator to achieve sufficient situational awareness for manipulation. One of the major factors affecting awareness is visual information, which benefits from quality live views during manipulation. The requirements for a quality view can vary dramatically over the course of the task, requiring a variety of positions and orientations. One solution to meet these requirements is to provide an adapting view from a camera on a drone. To reduce the cognitive load and effort associated with manual control of the drone camera, the drone viewpoint can be automated based on available environment and task information. However, while an automated drone has the potential to assist operators with the range of views required to support manipulation tasks, using a drone also introduces challenges. One challenge introduced by the mobility of the drone is how to select a viewpoint when there are multiple quality viewpoints spread across the workspace. Another challenge is how the viewpoint selection should compensate for drone flight characteristics, such as stability of the viewpoint and occlusions or collisions with objects in the environment. To address these challenges, we propose a method that combines human-toggling of automatically-generated drone viewpoints with an otherwise automated drone view that considers occlusions, collisions, and drone uncertainty.

\begin{figure}
\centering
\includegraphics[width=3.40in]{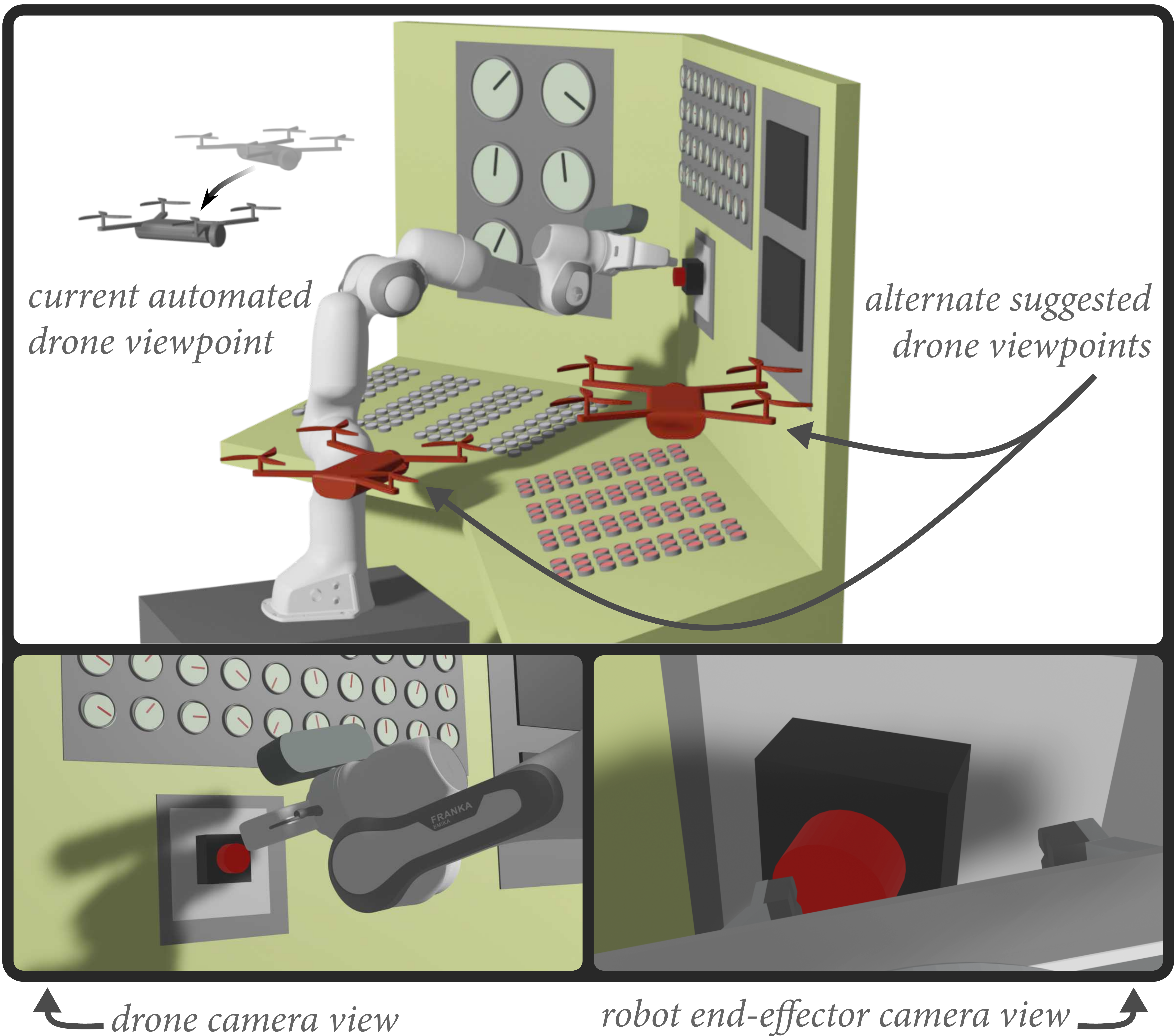}
\caption{\textit{Top:} A potential application of our method where an automated drone view supports a user teleoperating a robotic manipulator to press an e-stop button. Our method suggests alternate drone viewpoints (red) to which the operator can reposition the drone for additional perspective. \textit{Bottom:} The operator can use the adaptive drone view and additional visual information, such as a view from the end-effector camera, to complete the task.}
\label{fig:teaser}
\vspace{-15pt}
\end{figure}

In our implementation, an operator controls a manipulator through a remote interface and the drone automatically tracks the end-effector to provide an appropriate task view. Many existing works, summarized in Section \ref{sec:relatedwork}, have used drones to provide a dynamic view for applications such as search and rescue or cinematography. However, these existing methods have not focused on providing an automated drone view that is designed for cluttered manipulation environments. Manipulations often benefit from the perspective of multiple viewpoints of the task, for example to see specific parts of the manipulated object. However, particularly in cluttered environments, these viewpoints can require drastically different camera positions.
For example, when pressing a button on a panel, as shown in Figure \ref{fig:teaser}, two good viewpoints may be on opposite sides of the workspace. Our method leverages two simultaneous optimizations to allow operators to toggle between multiple viewpoints during manipulation tasks. The first optimization locally commands the drone pose. The second optimization globally queries the workspace, aggregating alternate viewpoints where the operators can reposition the drone, for example to get a different perspective of the manipulation task. When providing viewpoints in cluttered environments, the challenge of uncertainty in the drone's command tracking (i.e., small perturbations and oscillations from the desired pose) must also be addressed. Failing to account for uncertainty could lead to unexpected collisions or occluded views. 

In this work, we present a method to enable an automated drone viewpoint for telemanipulation tasks. Our method combines a set of nonlinear optimization objectives, which are defined based on environment modeling from a sensor mounted on the manipulator and empirically measured drone pose uncertainty. These objectives are used within a local-global optimization that both controls the drone and searches for and proposes alternate viewpoints to the operator. We provide an open-source implementation of our method which provides multiple live camera views from an RGB-D camera mounted to the manipulator and the drone on-board camera. Finally, we provide an initial validation of our method in an indoor, cluttered environment by performing manipulations of household and industrial objects.

%% file: 2-related.tex
\section{RELATED WORK} \label{sec:relatedwork}

\subsection{Methods for Camera Control}
Automatic camera control has been studied extensively in computer graphics to provide users with an effective view of a virtual environment (see \cite{christie2008camera} and \cite{galvane2015automatic} for surveys). Many of these existing methods rely on a fully unconstrained camera and a complete understanding of the virtual scene.
In real-world  scenarios, automatic camera control can be achieved using a robot (such as a drone) that dynamically and autonomously react to visual changes in the environment. In robotics, visual servoing \cite{chaumette2016visual} and active perception \cite{chen2011active, bajcsy2018revisiting} are control methods that are used to move a robot based on a stream of vision data. In visual servoing, features extracted from vision data are used to define a target pose for the robot and to determine how the robot should move \cite{hutchinson1996tutorial}. This technique has been used previously for autonomous drone control (e.g., in  \cite{xiao2017visual} and \cite{teuliere2011chasing}). 
In active perception, a robot (with an on-board camera) autonomously adapts its viewpoint to gain more information to complete tasks such as object recognition, manipulation, or reconstruction.  

We draw from the optimization-based techniques for camera control in computer graphics where visual properties are encoded as objectives, but include real-world drone considerations. Additionally, to address the specific case of telemanipulation, our method is designed to work with an approximate geometric model of the environment captured on site by sensors mounted on the manipulator. Our work shares parallels with 3D visual servoing, but unlike active perception where the viewpoint is adapted for a robot to autonomously complete a task, our method adapts the viewpoint to support human users completing manipulation tasks.


\subsection{Vision for Remote Robot Teleoperation}

During teleoperation, effective visual feedback from the remote environment is critical for the operator's situational awareness. A single static view of the remote environment is unlikely to provide all the visual information necessary for adequate awareness \cite{chen2007human}. While multiple static cameras can sometimes provide such information (e.g., in \cite{kent2020leveraging, sato2020cognitive}), the placement of the cameras is often strategically chosen for the scenario at hand, and slight variations in the task or the environment can make it difficult to complete the task. Additionally, at the time of robot deployment, environments may be inaccessible for setting up multiple static cameras. Hence, one or more dynamic cameras (i.e., cameras that can provide different views) can be valuable in the presence of varying task requirements. Manually controlling a dynamic camera can be effective in certain situations, such as first-person teleoperation of a humanoid robot (e.g., in \cite{pittman2014exploring, fritsche2015first} using head-tracking for camera control). However, in many telerobotics applications, manually controlling a dynamic camera introduces additional workload on the operator and reduces teleoperation performance. Prior work in laparoscopic surgery \cite{pandya2014review, wagner2021learning}, telemanipulation \cite{cabrera2021exploration, nicolis2018occlusion, rakita2018dynamic, rakita2019adapting}, and search and rescue \cite{shilleh2021best, xiao2021autonomous} has shown the effectiveness of employing \textit{autonomous} dynamic cameras. 

\subsection{Drone as Viewpoint Provider}

A dynamic camera can be moved using several mechanisms such as a gantry system \cite{rahnamaei2014automatic}, an articulated head \cite{schmaus2019knowledge}, a stationary robot arm \cite{rakita2019adapting}, a mobile robot arm \cite{maeyama2016view}, or a drone \cite{xiao2021autonomous}. These mechanisms provide different types of viewpoints to the user based on their range of motion and degrees of freedom. For example, the camera on an articulated head in \cite{schmaus2019knowledge} provides first-person pan-tilt views, whereas the camera on a drone in \cite{xiao2021autonomous} provides third-person, six degree-of-freedom views. Each mechanism for moving a dynamic camera has trade-offs. For example, a camera on a drone has high mobility, but uncertainty in its pose information due to flight dynamics that can result in unsafe, unstable, or occluded viewpoints. 
Drones have been used to provide views of teleoperated robots such as another drone \cite{temma2019third}, a construction machine \cite{kiribayashi2018design}, or a ground robot \cite{xiao2021autonomous, venzano2022motion}. However, these works have not considered cluttered manipulation environments where small errors in pose information can lead to collisions or occluded views. 

Drones are also widely used for autonomous aerial cinematography. Our work shares similar goals with drone cinematography such as visual objectives that define the quality of a viewpoint \cite{galvane2016automated,bonatti2018autonomous} and on-the-fly avoidance of occlusions and collisions \cite{nageli2017real}. However, visual objectives for telemanipulation need to be tailored to provide a quality viewpoint for manipulation tasks rather than to produce cinematic shots. Similarly, the occlusion and collision objectives can leverage the specific geometries present in manipulation scenarios. In many cinematography applications, events are scripted and the user is able to initially provide the system with specifications about desired viewpoints (for example, through keyframes in \cite{joubert2015interactive, gebhardt2016airways}). In contrast, the required views or regions of interest are not known beforehand in a number of scenarios where robots are deployed for telemanipulation. This prevents the user from providing any information ahead of time. Thus, telemanipulation scenarios also necessitate fast, online point-to-point optimization and preclude the use of offline trajectory optimization methods. 

\subsection{Summary}

Previous work has highlighted the opportunities provided by drones to provide viewpoints for robot operators. However, there remain a number of questions that need to be addressed when considering the specific scenario of telemanipulation in unknown cluttered environments. 

Similar to \cite{rakita2018dynamic, rakita2019adapting, dufek2021best}, our work adapts viewpoints to support human users completing manipulation tasks. Our work differs from previous work in drone camera control in three ways. Our work does not require a semantic understanding of the task or full representation of the workspace; instead we use a simplified geometric understanding of the manipulator, environment, and task (similar to \cite{brooks2001visual}) for selecting viewpoints. Second, we dynamically integrate user preferences by calculating and presenting the user with alternate viewpoints that can be used to reposition the drone. Finally, we draw from the work on drone pose uncertainty (see \cite{dadkhah2012survey} for a survey) and integrate the empirically measured drone pose error into our nonlinear optimization formulation for viewpoint selection.

%% file: 3-method.tex
\section{METHOD} \label{sec:method}

In our method, a drone is autonomously controlled to provide an operator with a dynamic camera view during manipulation tasks. In this section, we describe geometric cost functions that define a quality viewpoint while accounting for uncertainty in the drone pose and avoiding collisions with the environment. We also describe our solution to provide the operator with alternate drone viewpoints.

\subsection{Drone Characterization}
Viewpoint characteristics (e.g., occlusion) can be sensitive to small changes in position.
One challenge with using drones for dynamic viewpoints is compensating for errors in the drone pose caused by flight dynamics and disturbances. To address this challenge, we geometrically characterize the drone pose uncertainty and use this geometric characterization to inform our nonlinear optimization objectives.

Our premise is to empirically measure the maximum expected deviation of the drone pose. To do so, we use a series of command-tracking tasks (e.g., prolonged stationary pose tracking) to measure the discrepancies between the commanded and actual drone pose. The largest deviations in the drone state variables ($x$, $y$, $z$, $\theta$) are used to represent the uncertainty. Practically, drones can often control yaw (i.e., angle or $\theta$) with a relatively high degree of precision compared to Cartesian directions and we observed this effect on our platform too. Thus, we incorporate the uncertainty from the three Cartesian directions into the objectives involving geometric collisions and occlusions. Depending on the characteristics of a particular drone platform, it is also possible to add additional uncertain parameters as part of the geometric viewpoint objectives.


\begin{table*}[h]
\centering
    \caption{Symbols used in objectives.}
    \vspace{-0.05in}
    \begin{tabular}{p{0.1\linewidth} p{0.8\linewidth}}
    \toprule
    Symbol & Definition \\
    \midrule
    $f_{fov}$ & Field of view factor for scaling the distance from the drone to the end effector\\
    $\theta_{ref}$ & Reference angle for the perspective cost\\
    $\epsilon$ & Small constant used for the alternate view computation  \\
    $n$ & Number of planes\\
    $\theta_\alpha(\textbf{x})$ & Angle between the drone camera direction and the vector from the drone to the end effector\\
    $\theta_\beta(\textbf{x})$ & Angle between the vector from the end effector to the closest plane and the vector from the end effector to the drone\\
    $d_{dr\rightarrow plane_{k}}(\textbf{x})$ & Distance between the drone collision hull and the k$^{th}$ plane (negative distance when intersecting) \\
    $d_{dr\rightarrow mc}(\textbf{x})$ & Distance between the drone collision hull and the manipulator's collision hull\\
    $d_{dr\rightarrow ee}(\textbf{x})$ & Distance between the drone and the manipulator's end effector\\
    $d_{ee\rightarrow cp}(\textbf{x})$ & Distance between the end-effector and the closest point on a plane\\
    $d_{vh\rightarrow pl_{k}}(\textbf{x})$ & Distance between the viewing hull and the k$^{th}$ plane \\
    $d_{vh\rightarrow mo}(\textbf{x})$ & Distance between the viewing hull and the manipulator's occlusion hull.\\
    $d_{c\rightarrow new}(\textbf{x})$ & Distance between the current drone position and the proposed position \\
    $g_{\sigma}^P(.)$ & The higher-order Gaussian with order P and a standard deviation of $\sigma$: $g_{\sigma}^P(x) = e^{-(x^2/(2\sigma^2))^P}$ \\
    \bottomrule
    \end{tabular}
    \label{tab:obj_symb}
    \vspace{-0.05in}
\end{table*}

\subsection{Environment modeling}

To geometrically assess collisions and occlusions, the representation of the environment model needs to support rapid distance checking between shapes. 
In this work, we use planes as a simple geometric primitive to approximate the environment. Planar surfaces are pervasive in human-made environments (e.g., walls and tables), and there are many existing methods to recognize them from unstructured point cloud data.
We use these planes to evaluate both collisions and occlusions. Further, we assume that the manipulator will interact with objects on the planes. Thus, we approximate a region of interest for manipulation by calculating the point on the plane closest to the current position of the manipulator's end-effector. This approximation allows us to formulate objectives for viewing angle and distance that effectively support manipulation.

\begin{figure*}
\centering
\includegraphics[width=\textwidth]{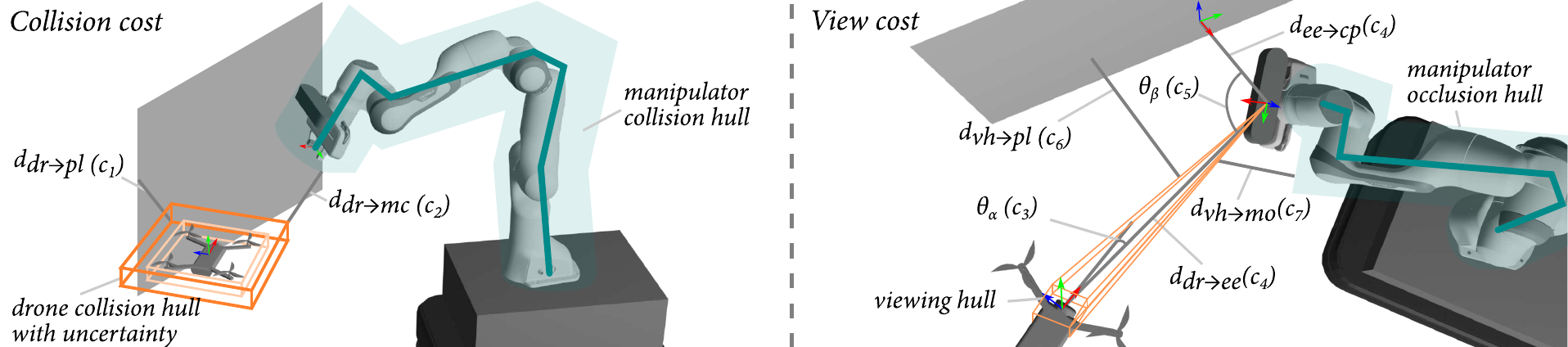}
\caption{Illustration of geometric quantities used in the objectives. The objective function that uses the geometric quantity is denoted in parantheses.}
\label{fig:objectives}
\end{figure*}

We model the drone and manipulator by enveloping their geometries with simple primitives to obtain convex hulls that are used to approximate collision, viewing, and occlusion regions.
The drone collision hull is modeled as a cuboid encapsulating the drone geometry. Each dimension of the cuboid is extended by the empirically-measured drone uncertainty. The drone viewing hull is modeled as a convex hull connecting a cuboid representing the drone camera's position and the manipulator's end effector position. The cuboid is centered on the drone camera position and has the dimensions of the drone pose uncertainty (see Figure \ref{fig:objectives}). The manipulator collision hull is represented by a polyline connecting each of the joints from the base to the end-effector with a constant width based on the width of the robot. Finally, the manipulator occlusion hull is represented with the same structure as the collision hull but without the end effector as part of the polyline. 

\subsection{Drone pose optimization}

The commanded drone pose is calculated online using a weighted-sum nonlinear optimization:
\begin{equation}
        \argmin_{\textbf{x}}\sum\limits_{i}w_i c_i(\textbf{x})
    \end{equation}
where $c_i$ are weighted objective terms (introduced below) with corresponding weights $w_i$ and $\textbf{x}$ contains the drone state variables (e.g., $x$, $y$, $z$, $\theta$).
The weighted objectives can be divided into three categories: collision objectives, view objectives, and motion-limiting objectives. To improve performance, the cost function at each iteration is optimized over a small rectangular search space from the last commanded pose. Symbols used in the objectives are defined in Table \ref{tab:obj_symb}.

\subsubsection{Collision objectives}
These objectives steer the drone away from collisions with the environment and the manipulator.
\begin{itemize}[leftmargin=*]
    \item \textit{Environment collision:} The drone should stay away from planes.
    \begin{equation}
        c_1(\textbf{x}) = \sum_{k=1}^{n} g_{\sigma}^P(d_{dr\rightarrow pl_{k}}(\textbf{x}))_+
    \end{equation} 
    \item \textit{Manipulator collision:} The drone should stay away from the manipulator.
    \begin{equation}
        c_2(\textbf{x}) = g_{\sigma}^P(d_{dr\rightarrow mc}(\textbf{x}))_+
    \end{equation}
\end{itemize}

\subsubsection{View objectives}
These objectives provide a view that supports the operator in doing the current task. The view cost balances viewing the manipulator end effector (the element interacting with the environment) and providing enough context to solve the task. This formulation does not rely on object recognition nor inferring the task goal. We instead use geometric relations to generalize to different tasks and environments.
\begin{itemize}[leftmargin=*]
    \item \textit{Visual target:} The camera should be centered on the manipulator end effector.
    \begin{equation}
        c_3(\textbf{x})=(\theta_\alpha(\textbf{x}))^2
    \end{equation}
    \item \textit{Distance to visual target:} When the manipulator is close to a plane, it might indicate the operator's intention to interact with an object on this plane. In this case, the drone should move closer to the manipulator.
    \begin{equation}
        c_4(\textbf{x})=(d_{dr\rightarrow ee}(\textbf{x})-f_{fov}\cdot d_{ee\rightarrow cp}(\textbf{x}) -d_{min})^2
    \end{equation}
    \item \textit{Perspective angle:} The drone should maintain a specified angle with the plane to provide perspective of any objects on the plane.
    \begin{equation}
        c_5(\textbf{x})=(\theta_\beta(\textbf{x})-\theta_{ref})^2
    \end{equation}
    \item \textit{Environment occlusion:} The drone's viewing hull should avoid intersection with environment planes.
    \begin{equation}
        c_6(\textbf{x})=\sum\limits_{k=1}^{n} ((d_{vh\rightarrow pl_{k}}(\textbf{x}))_-)^2
    \end{equation}
    \item \textit{Manipulator occlusion:} The drone's viewing hull should avoid intersection with the manipulator's occlusion hull (a polyline that excludes the end effector).
    \begin{equation}
        c_7(\textbf{x})=((d_{vh\rightarrow mo}(\textbf{x}))_-)^2
    \end{equation}
\end{itemize}

\subsubsection{Motion-limiting objectives}
This objective ensures that the current commanded position is within reach of the drone.
\begin{itemize}[leftmargin=*]
    \item \textit{Distance current drone position goal:} The commanded position should be close to the current position.
    \begin{equation}
        c_8(\textbf{x}) = (d_{c\rightarrow new}(\textbf{x}))^6
    \end{equation}
\end{itemize}

\begin{figure*}[b]
\centering
\vspace{-0.15in}
\includegraphics[width=\textwidth]{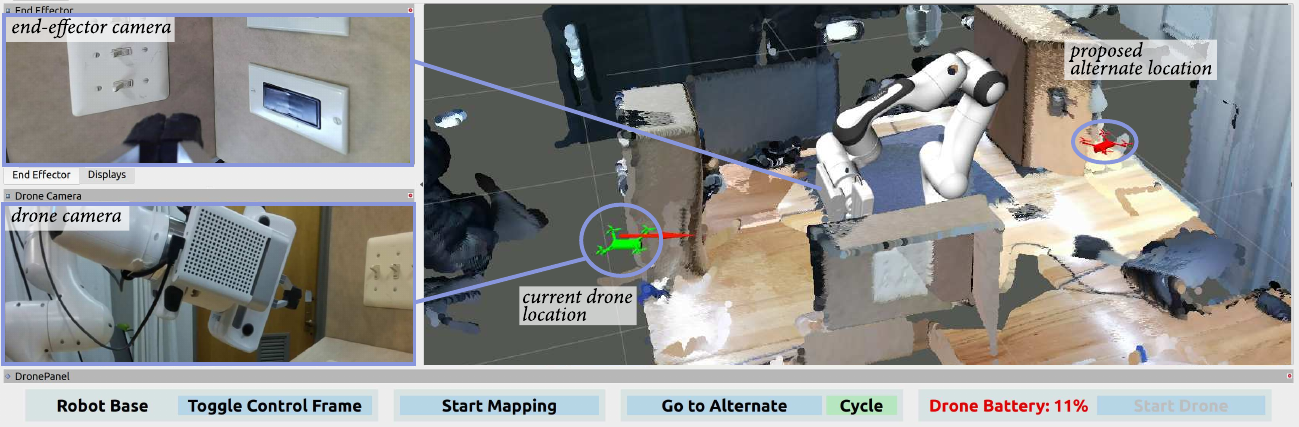}
\caption{Annotated interface used for teleoperation (the aspect ratio has been updated for clarity in the paper).}
\label{fig:interface}
\end{figure*}

\subsection{Alternate Viewpoint Generation and Selection}
There are a variety of situations where the operator may desire a different view than what the drone is currently providing. For example, a view from a different angle in the workspace may provide additional perspective or help to gather additional task-specific information (e.g., reading a gauge on one side of a valve). There may also be instances where the online drone viewpoint optimization converges to a local minimum that is insufficient to complete the task. 

In these cases, we provide the operator with a small number of quality alternate views for the drone camera. This is accomplished through a second viewpoint optimization that searches globally for alternative views that minimize the drone cost function. Specifically, this optimization uses only the previously defined view and collision objectives (i.e., no motion-limiting objectives). This optimization is seeded using a random starting drone pose and searches the full workspace at each iteration. Additionally, we add one objective to this optimization which penalizes viewpoints that are similar to the current drone viewpoint:
\begin{equation}
        c_9 = 1/(\epsilon + d_{c\rightarrow new}(\textbf{x})) 
    \end{equation}

The viewpoints from this optimization are clustered and a predefined number of top candidates are stored to maintain a diversity of viewpoints. Stored alternate viewpoints expire after a short amount of time to encourage new viewpoints as the manipulator changes pose. During operation, the user is able to cycle through the alternate viewpoints using a simple toggle. When the user identifies a preferential viewpoint, they can signal the drone to transition to the alternate viewpoint.


    



%% file: 4-implementation.tex
\section{IMPLEMENTATION} \label{sec:implementation}

To test our method for automated drone viewpoint assistance, we developed a manipulator-drone system with a multiple-view remote interface. In this section, we describe our apparatus, optimization details, and interface design. All system-specific implementation quantities (e.g., drone uncertainty, optimization weights) appear in Table \ref{table:constant}. Our code is available online.\footnote{\url{https://github.com/emmanuel-senft/drone-panda}}

\subsection{Apparatus}
The setup consists of a Franka Emika robot controlled in joint position. The desired joint positions are calculated by a nonlinear inverse kinematics optimization that balances matching the desired end-effector pose, limiting joint velocity, and avoiding robot joint limits. The robot is equipped with a gripper, ATI force-torque sensor, and Kinect Azure RGB-D camera mounted to the distal link with a known transform. The drone camera is provided by a Ryze Tello drone. The drone position is captured used a Optitrack motion capture system and a \emph{proportional-derivative} (PD) controller is used to command the drone velocity through a python wrapper based on the official Tello SDK.\footnote{\url{https://github.com/damiafuentes/DJITelloPy}}

The different components of the system communicate using ROS \cite{quigley2009ros}. The environment is mapped from the Kinect's point cloud using RTAB-Map \cite{labbe2019rtab} and is acquired by running an automated routine that controls the robot-mounted camera to create an environment map. Planes are extracted from the resulting point cloud using Point Cloud Library's \cite{Rusu_ICRA2011_PCL} plane segmentation. The resulting plane locations and inlier points are further processed to determine plane boundaries (i.e., edges) and non-rectangular surfaces are subdivided into smaller planes as necessary using a recursive routine.

\begin{table}[]
\centering
  \caption{Constant values in implementation and optimization. }
  \label{table:constant}
  \vspace{-0.05in}
  \setlength\tabcolsep{4pt}
  \begin{tabular}{ll|ll|ll|ll}
    \toprule
    \multicolumn{8}{c}{Characterization}\\
    \cmidrule(r){1-8}
    \textit{C} & \textit{Value} & \textit{C} & \textit{Value} & \textit{C} & \textit{Value} & \textit{C} & \textit{Value} \\
    $\Delta p_x$ & 0.05 m & $\Delta p_y$ & 0.05 m & $\Delta p_z$ & 0.02 m & robot size & 0.1m\\
    \midrule
    \multicolumn{8}{c}{Optimization weights}\\
    \cmidrule(r){1-8}
    $w_1$ & 50 & $w_2$ & 50 & 
    $w_3$ & 10 & $w_4$ & 10\\
    $w_5$ & 5 & $w_6$ & 5000 & 
    $w_7$ & 100 & $w_8$ & 1000\\
    \midrule
    \multicolumn{8}{c}{Optimization constants}\\
    \cmidrule(r){1-8}
    $f_{fov}$ & 2 & $P$ & 3 &
    $\theta_{ref} $ & 4$\pi$/3 rad & $\sigma$ & 0.1m\\
    \midrule
    \multicolumn{8}{c}{Alternate viewpoint constants}\\
    \cmidrule(r){1-8}
    $\epsilon$ & 0.1 & $w_9$ & 1\\
\bottomrule
  \end{tabular}
\end{table}

\subsection{Optimization}
Our implementation consists of three optimization routines that determine the drone pose, manipulator pose, and alternate drone viewpoints. Each optimization was written in RUST and is solved using the PANOC method of OpEn library \cite{open2020}. The search space, relative to the current drone pose, was bounded to $\pm 1$  cm for each dimension of the drone position and $\pm 0.1$ rad for the drone orientation, and $\theta_{ref}$ and $w$ were set empirically. A C++ wrapper connects the RUST code to ROS. The drone and manipulator pose optimizations run at 100 Hz whereas the alternate viewpoint optimization runs at 5 Hz. All of the geometric collisions and occlusions are computed using the ncollide3d\footnote{\url{https://ncollide.org/}} library. When the operator selects an alternative viewpoint, we freeze the manipulator control and use a simple path planning procedure where the drone ascends to the maximum height, moves to above the selected alternative viewpoint, and descends. 

\subsection{Interface and Controls}
For telemanipulation, we developed a multi-view interface within RViz (as shown in Figure \ref{fig:interface}) which combines the drone camera view, end-effector camera view, and a simulated view. The simulated view gives a global perspective that consists of the environment point cloud, manipulator, current drone position, and the alternate proposed drone position. The interface also contains a series of widgets at the bottom of the interface, such as toggling the control frame, toggling the alternate view, commanding the drone to move to an alternate view, and the drone battery level. The available control frames are the drone camera frame, the end-effector frame, the manipulator base frame, and the simulated view camera frame. The simulated view also contains visualization of the end effector's applied torque to assist in tasks involving physical interaction, such as turning valves.

The operator uses a gamepad controller to control the 6D pose of the manipulator and to provide high-level commands to the drone such as \emph{takeoff}, \emph{land}, and \emph{move to alternate view}.

%% file: 5-case_study.tex
\section{PRELIMINARY STUDY} \label{sec:casestudy}

\begin{figure}
\centering
\includegraphics[width=3.30in]{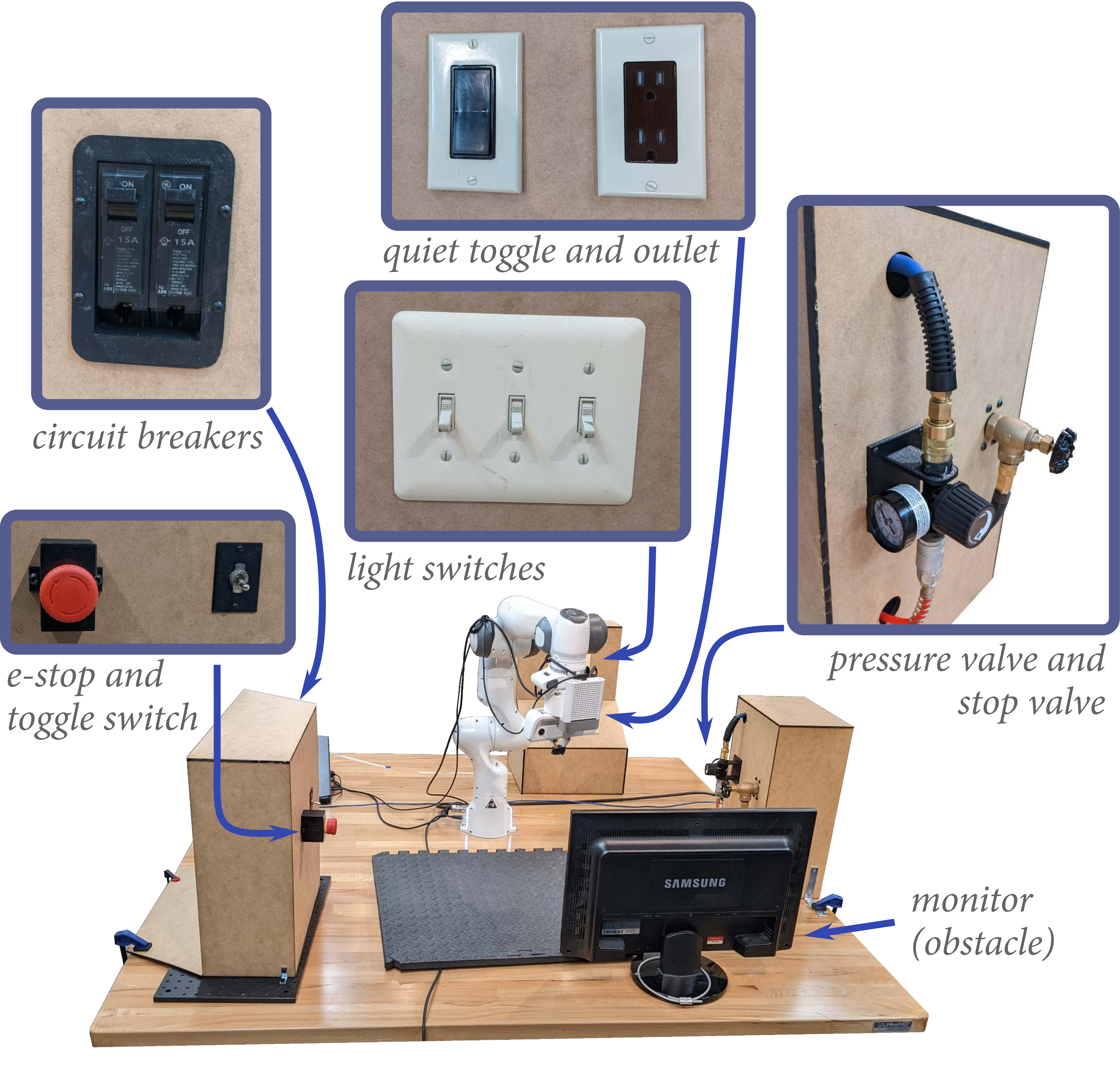}
\caption{Workspace and tasks used in the validation study.}
\label{fig:workspace}
\vspace{-0.2in}
\end{figure}

\begin{figure*}
\centering
\includegraphics[width=\textwidth]{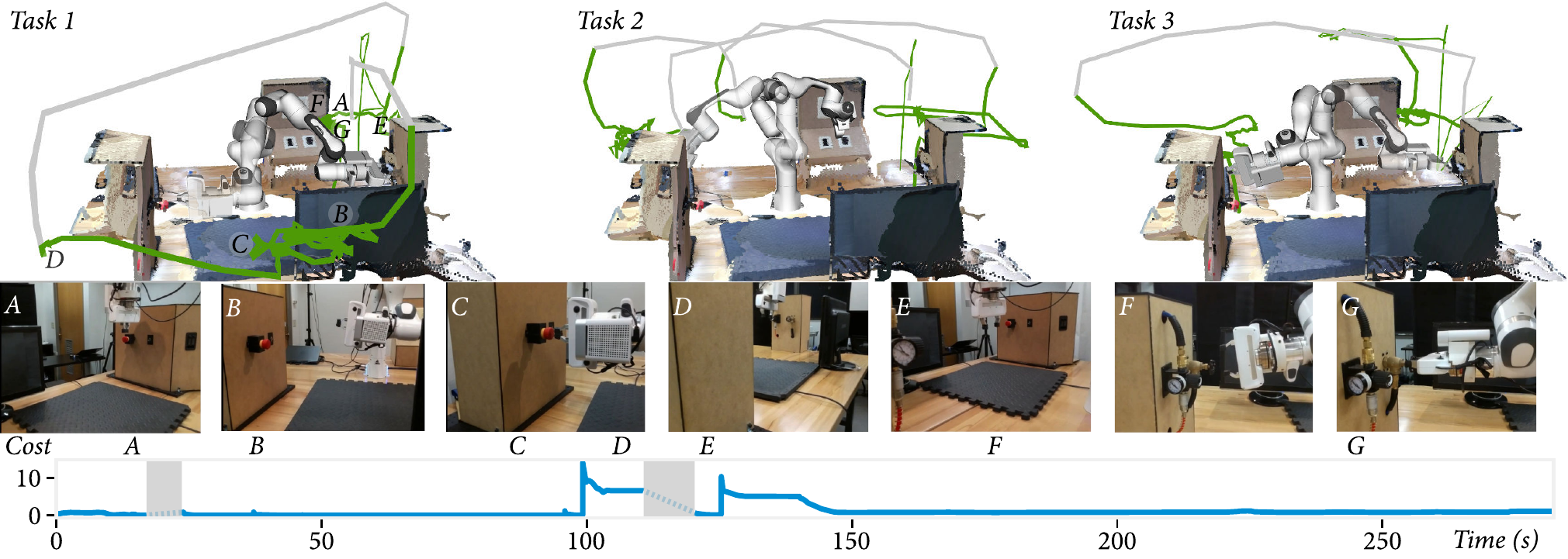}
\caption{\textit{Top:} Drone trajectory for each of the three tasks. Gray indicates transitions between alternate viewpoints. \textit{Middle:} Screenshots of the drone view at different times and viewpoints in Task 1 as a case study. \textit{Bottom:} Evolution of the cost (view and collision) for Task 1.}
\label{fig:results}
\vspace{-0.2in}
\end{figure*}

To demonstrate our method, we designed a series of common household and industrial manipulation tasks that involved operating an e-stop button, switches, circuit breakers, and valves. The environment was setup as shown in Figure \ref{fig:workspace}. To assess the behavior of our system in these tasks, we carried out a preliminary validation study, involving a single expert operator. In the future, we plan to conduct user studies for a more comprehensive and multifaceted evaluation.

\subsection{Tasks}
We designed three tasks with multiple manipulations spread around the workspace. In each task, the manipulator was controlled by an expert operator located in a different room than the manipulator and drone setup.

\begin{itemize}[leftmargin=*]
    \item \emph{Task 1:} Press the e-stop button and turn the stop valve to a closed configuration.
    \item \emph{Task 2:} Turn off both circuit breakers, turn off the three light switches, and turn off the quiet light switch.
    \item \emph{Task 3:} Turn the pressure valve to 20 psi and turn off the toggle switch.
\end{itemize}

Our primary goal of this preliminary validation study was to show examples of drone viewpoints and alternate views that our system proposes over the course of manipulation tasks in cluttered environments. A secondary goal was to show that our proposed nonlinear optimization was able to avoid occlusions and collisions in live tests. Periods of occlusion (i.e., when there was a partial or no view of the end effector) and collisions for the drone were coded by two of the authors. 
Transitions between viewpoints (i.e., alternate viewpoints) were not considered in determining occlusions.

\subsection{Results and Observations}
Figure \ref{fig:results} shows the trajectory of the drone for each task. Example drone viewpoints are also featured in the supplemental video.\footnote{A full recording of each task is available at \url{https://youtube.com/playlist?list=PL76JjYsZbSTkieqgS4C1LiYVDvLOgpXv1}} We also provide additional detail for Task 1 as a case study, including example views (middle of Figure \ref{fig:results}), the evolution of the view and collision cost function (bottom of Figure \ref{fig:results}), and a snapshot of the cost map (view and collision) in a state similar to Figure \ref{fig:results}.D (see Figure \ref{fig:heat}). No collisions were observed during the execution of the tasks. Notable observations during each task are described below.

\textit{Task 1} had a duration of 4:36 (four minutes and 36 seconds). No periods of occlusion were noted during this task. A notable observation was that, as the manipulator approached the stop valve, the drone view was distant and appeared to be in a local minimum. To obtain a more informative viewpoint, the operator repositioned the drone to an alternate side view of the valve (viewpoint E).

\textit{Task 2} had a duration of 4:58. There was a one second occlusion as the end effector rotated to approach the circuit breakers, which prompted the operator to reposition the drone to a side view. There was a three second occlusion of the end effector when approaching the light switches that was automatically resolved by the drone controller.

\textit{Task 3} had a duration of 3:53. There was an occlusion for five seconds as the manipulator approached the valve. The drone view was higher than desirable and appeared to be in a local minimum. The operator repositioned the drone for a closer, level view of the end effector. There was another occlusion of the gripper for less than one second that the drone resolved automatically. When the manipulator interacted with the toggle switch, the drone maintained a viewpoint that was nearly orthogonal to avoid collisions, which caused the switch to become briefly occluded for less than one second while the end effector remained visible.




%% file: 6-discussion.tex
\section{DISCUSSION} \label{discussion}
\begin{figure}[!b]
\centering
\vspace{-0.2in}
\includegraphics[width=0.95\columnwidth]{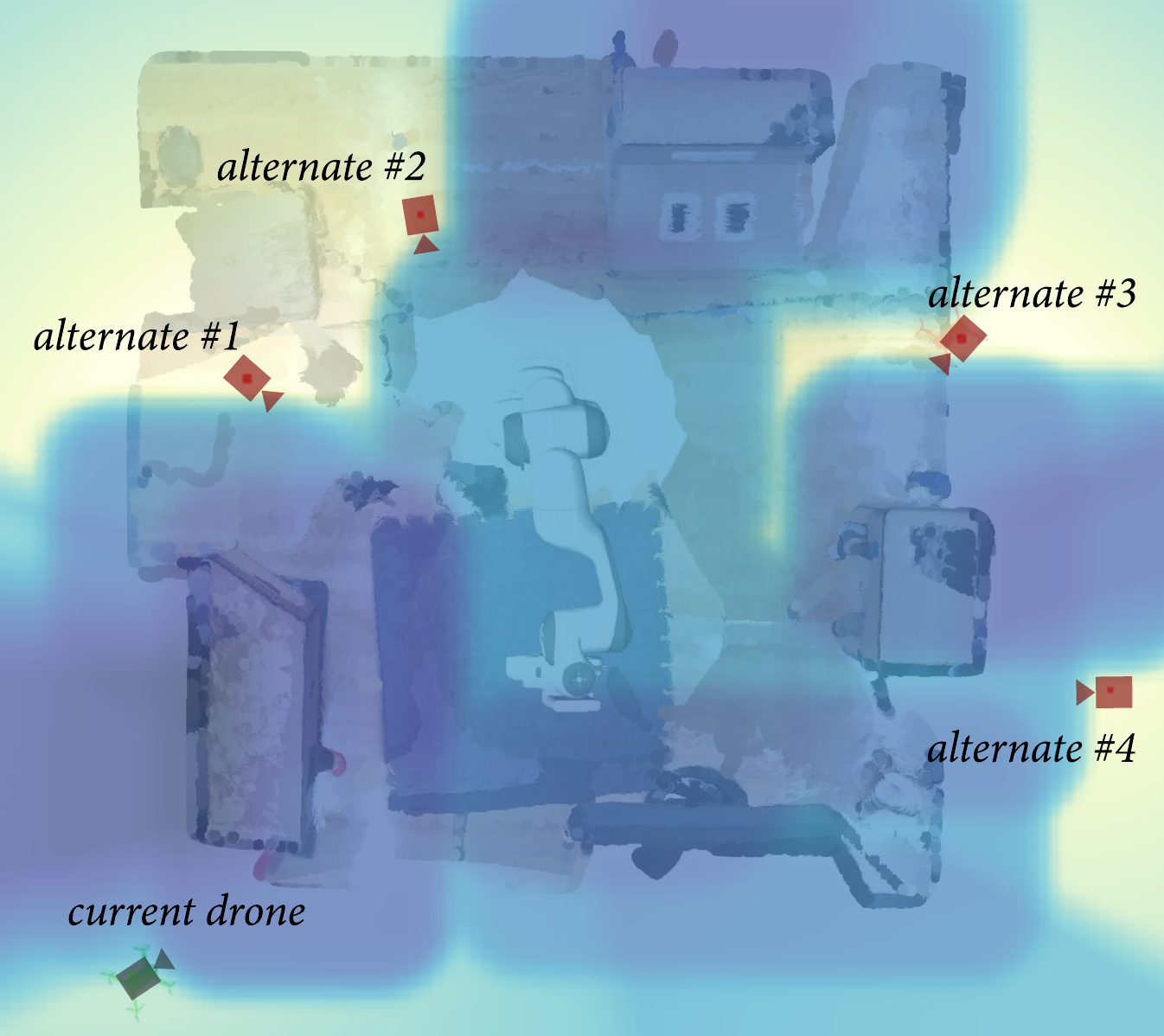}
\caption{Example of the cost map (view and collision) in a state similar to Figure \ref{fig:results}.D. Colors are plotted on a log scale and lighter colors indicate lower cost. The current position is a local minimum and alternative viewpoints are proposed. Figure \ref{fig:results}.E corresponds to a relocation to alternate \#3.}
\label{fig:heat}
\end{figure}

Our method was able to generally provide an unoccluded view of the manipulator. Observed cases of occlusion were brief and were resolved either by the autonomous viewpoint adaption or, in cases of local minima, by the operator switching to an alternate viewpoint. We believe that these preliminary results demonstrate the promise of our proposed contributions. The local-global optimization is able to generate a range of unoccluded alternate viewpoints, as shown in Figure \ref{fig:heat}, that can offer additional visual task information. The lack of collisions and minimal occlusion observed in our validation study illustrate the value of the geometric objectives conditioned on the drone uncertainty.

\textit{Limitations}---
Our method has some inherent limitations. First, our method relies on a static map of the environment and does not consider new or dynamic obstacles. Our map also comes from a single scan from the manipulator robot, which does not model all possible obstacles in the environment. In the future, we would like to supplement the robot map with online mapping from the drone that can both update changing parts of the map and model areas missed by the initial robot scan. Our method assumes environment features can be modeled as planes. While many engineered environments are flat, there are certain environments (e.g., curved surfaces, free-standing objects) that may not work well with our objective structure. All drone viewpoint heuristics were modeled as costs, rather than constraints. While we did not note any collisions during our preliminary evaluation, a constraint-based formulation may be appropriate in other circumstances. We geometrically bounded the drone uncertainty, however in some environments, this may be too conservative and a probabilistic model of uncertainty may be more appropriate. In some cases, the user may desire a specific viewpoint based on task semantics (e.g., reading a gauge). In the future, we would like to supplement our method with additional user controls.

Our implementation and preliminary validation had limitations we plan to address in future work. First, our drone had only four degrees of freedom, which limited the viewpoints that the optimization could consider. In the future, we would like to explore drones with a pan-tilt camera, which can provide additional views, such as overhead. Our implementation also relied on external drone state estimation via motion capture. Future work could explore how our method works with other techniques for localization. Finally, our method and system should also be evaluated in a user study with a wider range of tasks.


\textit{Conclusion}---
We presented a method for automated drone viewpoints during robot telemanipulation. We developed an implementation and showed through a preliminary validation that it can provide collision-free unoccluded views during common manipulation tasks.